\documentclass[journal]{IEEEtran}

\ifCLASSINFOpdf
\else
\fi

\usepackage[utf8]{inputenc} 
\usepackage[T1]{fontenc}    
\usepackage{hyperref}       
\usepackage{url}            
\usepackage{booktabs}       
\usepackage{amsfonts}       
\usepackage{nicefrac}       
\usepackage{microtype}      

\usepackage{graphicx}
\usepackage{subfigure}
\usepackage{multirow}
\usepackage{amssymb}
\usepackage{amsmath}
\usepackage{booktabs}
\usepackage{algorithm}
\usepackage{algorithmic}
\usepackage{color}
\usepackage{wrapfig}

\begin{document}
\title{Filter Sketch for Network Pruning}

\author{Mingbao Lin,
        Liujuan Cao,
        Shaojie Li, 
        Qixiang Ye,~\IEEEmembership{Senior Member,~IEEE}, 
        Yonghong Tian,~\IEEEmembership{Senior Member,~IEEE},
        Jianzhuang Liu,~\IEEEmembership{Senior Member,~IEEE},
        Qi Tian~\IEEEmembership{Fellow,~IEEE},
        Rongrong Ji,~\IEEEmembership{Senior Member,~IEEE},
\IEEEcompsocitemizethanks{

\IEEEcompsocthanksitem M. Lin, S. Li and R. Ji are with the Media Analytics and Computing Laboratory, Department of Artificial Intelligence, School of Informatics, Xiamen University, Xiamen 361005, China.\protect
\IEEEcompsocthanksitem L. Cao (Corresponding Author) is with the Fujian Key Laboratory of Sensing and Computing for Smart City, Computer Science Department, School of Informatics, Xiamen University, Xiamen 361005, China (e-mail: caoliujuan@xmu.edu.cn)
\IEEEcompsocthanksitem R. Ji is also with Institute of Artificial Intelligence, Xiamen University, Xiamen 361005, China.\protect
\IEEEcompsocthanksitem Q. Ye is with School of Electronic, Electrical and Communication Engineering, University of Chinese Academy of Sciences, Beijing 101408, China.\protect 
\IEEEcompsocthanksitem Y. Tian is with the School of Electronics Engineering and Computer Science, Peking University, Beijing 100871, China.\protect 
\IEEEcompsocthanksitem J. Liu is with the Noah's Ark Lab, Huawei Technologies Co. Ltd., Shenzhen 518129, China. \protect
\IEEEcompsocthanksitem Q. Tian is with Cloud BU, Huawei Technologies Co. Ltd., Shenzhen 518129, China.\protect 
}
\thanks{Manuscript received April 19, 2005; revised August 26, 2015.}}

\markboth{IEEE TRANSACTIONS ON NEURAL NETWORKS AND LEARNING SYSTEMS (Accepted)}%
{Shell \MakeLowercase{\textit{et al.}}: Bare Demo of IEEEtran.cls for IEEE Journals}

\maketitle

\begin{abstract}
We propose a novel network pruning approach by information preserving of pre-trained network weights (filters). Network pruning with the information preserving is formulated as a matrix sketch problem, which is efficiently solved by the off-the-shelf Frequent Direction method. Our approach, referred to as FilterSketch, encodes the second-order information of pre-trained weights, which enables the representation capacity of pruned networks to be recovered with a simple fine-tuning procedure. FilterSketch requires neither training from scratch nor data-driven iterative optimization, leading to a several-orders-of-magnitude reduction of time cost in the optimization of pruning. Experiments on CIFAR-10 show that FilterSketch reduces 63.3\% of FLOPs and prunes 59.9\% of network parameters with negligible accuracy cost for ResNet-110. On ILSVRC-2012, it reduces 45.5\% of FLOPs and removes 43.0\% of parameters with only 0.69\% accuracy drop for ResNet-50. Our code and pruned models can be found at \url{https://github.com/lmbxmu/FilterSketch}.
\end{abstract}

\begin{IEEEkeywords}
network pruning, sketch, filter pruning, structured pruning, network compression \& acceleration, information preserving
\end{IEEEkeywords}

\IEEEpeerreviewmaketitle

\section{Introduction}
\IEEEPARstart{D}{eep} convolutional neural networks (CNNs) typically result in significant memory requirement and computational cost, hindering their deployment on front-end systems of limited storage and computational power. Consequently, there is a growing need for reduction of model size by parameter quantization \cite{cheng2017quantized,wang2020unsupervised}, low-rank decomposition \cite{kim2015compression,hayashi2019exploring}, and network pruning \cite{chen2020dynamical,lin2019toward}. 
Early pruning works \cite{lecun1990optimal,han2015learning} use unstructured methods to obtain irregular sparsity of filters. Recent works pay more attention to structured pruning \cite{singh2019play,zhao2019variational,lin2019towards}, which pursues simultaneously reducing model size and improving computational efficiency, facilitating model deployment on general-purpose hardware and/or usage of basic linear algebra subprograms (BLAS) libraries.

%

Existing structured pruning approaches can be classified into three categories: 
(1) Regularization-based pruning, which introduces sparse constraint \cite{liu2017learning,huang2018data,zhao2019variational} and mask scheme \cite{lin2019towards} in the training process. Despite its simplicity, this kind of approaches usually requires to train from scratch and therefore is computationally expensive. 
(2) Property-based pruning, which picks up a specific property of a pre-trained network, \emph{e.g.}, $\ell_1$-norm \cite{li2017pruning} and/or ratio of activations \cite{hu2016network}, and simply removes filters with less importance. However, many of these approaches require to recursively prune the filters of each single pre-trained layer and fine-tune, which is very costly.
(3) Reconstruction-based pruning \cite{he2017channel,luo2017thinet}, which imposes relaxed constraints to the optimization procedure. Nevertheless, the optimization procedure in each layer is typically data-driven and/or iterative \cite{he2017channel,luo2017thinet}, which brings a heavy optimization burden.

In this paper, we propose the FilterSketch approach, which, by encoding the second-order information of pre-trained network weights, provides a new perspective for deep CNN compression. 
FilterSketch is inspired by the fact that preserving the second-order covariance of a matrix is equal to maximizing the correlation of multi-variate data \cite{sun2017svdnet}. The representation of the second-order information has been demonstrated to be effective in many other tasks \cite{kim2018bilinear,zhang2019discriminative,ren2019heterogeneous}, and for the first time, we apply it to network pruning in this paper.

\begin{figure*}[!t]
\begin{center}
\includegraphics[height=0.35\linewidth]{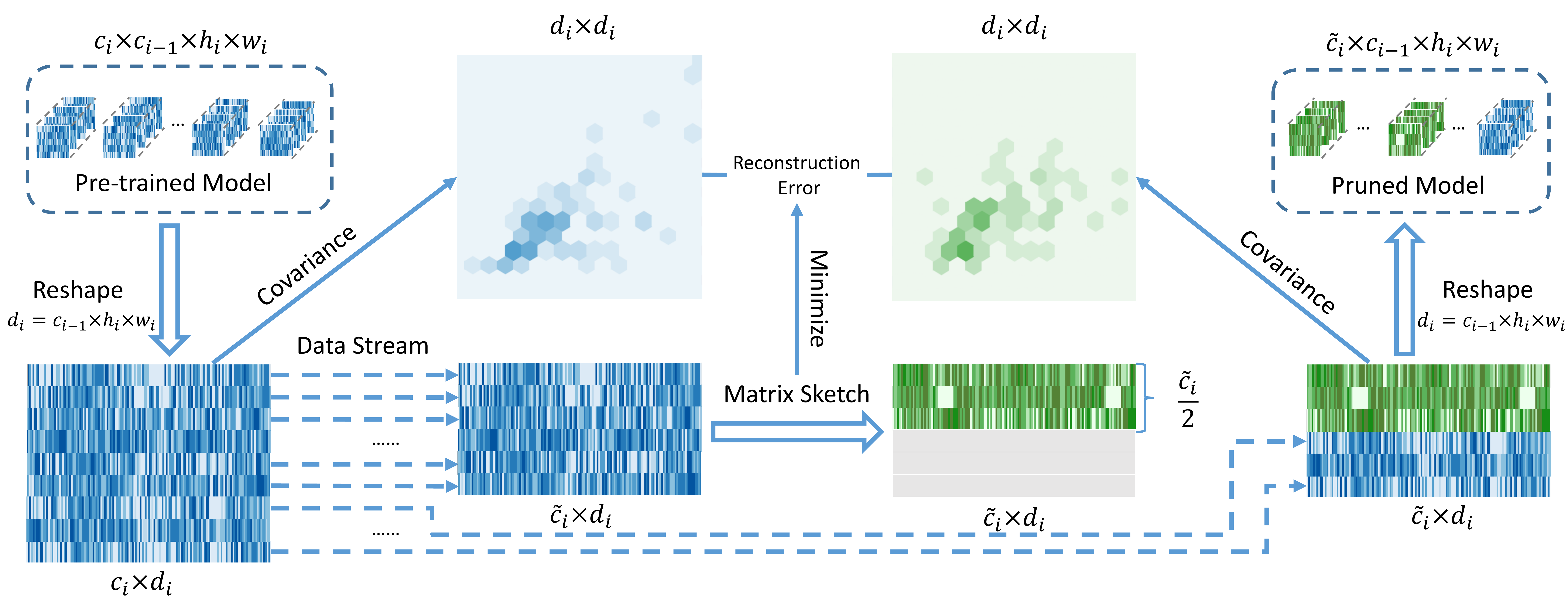}
\end{center}
\caption{\label{framework}
Framework of FilterSketch. The upper part displays the second-order covariance approximation between the pre-trained model and the pruned model at the $i$-th layer. The lower part shows the approximation is achieved effectively and efficiently by the stream data based matrix sketch \cite{liberty2013simple}. 
}
\end{figure*}

Instead of simply discarding the unimportant filters, FilterSketch preserves the second-order information of the pre-trained model in the pruned model as shown in Fig.\,\ref{framework}. For each layer in the pre-trained model, FilterSketch learns a set of new parameters for the pruned model which maintains the second-order covariance of the pre-trained model. The new group of sketched parameters then serves as a warm-up for fine-tuning the pruned network. The warm-up provides an excellent ability to recover the model performance.
%

We show that preserving the second-order information can be approximated as a matrix sketch problem, which can then be efficiently solved by the off-the-shelf Frequent Direction method \cite{liberty2013simple}, leading to a several-orders-of-magnitude reduction of optimization time in pruning. FilterSketch thus involves neither complex regularization to restart retraining nor data-driven iterative optimization to approximate the covariance information of the pre-trained model.

\section{Related Work}\label{related_work}
Unstructured pruning and structured pruning are two major lines of methods for network model compression. In a broader view, \emph{e.g.}, parameter quantization and low-rank decomposition can be integrated with network pruning to achieve higher compression and speedup. We give a brief discussion over some related works in the following and refer readers to the survey paper~\cite{vadera2020methods} for a more detailed overview.

\textbf{Unstructured Pruning}. Pruning a neural network to a reasonable smaller size and for a better generalization, has long been investigated. 
As a pioneer work, the second-order Taylor expansion~\cite{lecun1990optimal} is utilized to select less important parameters for deletion. Sum~\emph{et al.}~\cite{sum1999kalman} introduced an extended Kalman filter (EKF) training to measure the importance of a weight in a network. Given the recursive Bayesian property of EKF, they further considerd the sensitivity of a posterior probability as a measure of weight importance, which is then applied to prune in a nonstationary environment~\cite{sum1999adaptive} or recurrent neural networks~\cite{sum1998extended}.

Han~\emph{et al.}~\cite{han2015learning} introduced an iterative weight pruning method by fine-tuning with a strong $\mathit{l}_2$ regularization and discarding the small weights with values below a threshold.
Group sparsity based regularization of network parameters~\cite{alvarez2016learning} is leveraged to penalize unimportant parameters.
Further, \cite{dong2017learning} prunes parameters based on the second-order derivatives of a layer-wise error function, while~\cite{liu2018frequency} implements CNNs in the frequency domain and apply 2-D DCT transformation to sparsify the coefficients for spatial redundancy removal. 
The lottery ticket hypothesis \cite{frankle2019lottery} sets the weights below a threshold to zero, rewinds the rest of the weights to their initial configuration, and then retrains the network from this configuration. 

Though progress has been made, unstructured pruning requires specialized hardware or software supports to speed up inference. It has limited applications on general-purpose hardware or BLAS libraries in practice, due to the irregular sparsity in weight tensors.

\textbf{Structured Pruning}.
Compared to unstructured pruning, structured pruning does not have limitations on specialized hardware or software since the entire filters are removed, and thereby it is more favorable in accelerating CNNs.

To this end, regularization-based pruning techniques require a joint-retraining from scratch to derive the values of filters such that they can be made sufficiently small. To that effect, \cite{liu2017learning,zhao2019variational} impose a sparse property on the scaling factor of the batch normalization layer with the deterministic $\ell_1$-norm or dynamical distribution of channel saliency. After re-training, the channels below a threshold are discarded correspondingly. 
Huang \emph{et al}. ~\cite{huang2018data} proposed a data-driven sparse structure selection by introducing scaling factors to scale the outputs of the pruned structure and added the sparsity constraint on these scaling factors.
Lin \emph{et al}.~\cite{lin2019towards} proposed to minimize an objective function with $l_1$-regularization on a soft mask via a generative adversarial learning and adopted the knowledge distillation for optimization.

Property-based pruning tries to figure out a discriminative property of pre-trained CNN models and discards filters of less importance. 
Hu~\emph{et al.}~\cite{hu2016network} utilized the abundant zero activations in a large network and iteratively prunes filters with a higher percentage of zero outputs in a layer-wise fashion. 
Li~\emph{et al.}~\cite{li2017pruning} used the sum of absolute values of filters as a metric to measure the importance of filters, and assumes filters with smaller values are less informative and thus should be pruned first. In \cite{yu2018nisp}, the importance scores of the final responses are propagated to every filter in the network and the CNN is pruned by removing the filter with the least importance. Polyak~\emph{et al}.~\cite{polyak2015channel} considered the contribution variance of each channel and removed filters that have low-variance outputs. It shows a good ability to accelerate the deep face networks. Centripetal SGD~\cite{ding2019centripetal} constrains filters within the same cluster to move towards a center, and thus removes the identical filters without the necessity of fine-tuning. EigenDamage~\cite{wang2019eigendamage} decorrelates filter weight on top of the Kronecker-factored eigenbasis to enable weights to be approximately independent, meanwhile it allows a global ranking of filter importance in Hessian-based pruning. He \emph{et al}.~\cite{he2019filter} proposed to calculate the geometric median in each layer, and the filters closest to this are pruned.

\begin{figure*}[!t]
\begin{center}
\includegraphics[height=0.22\linewidth]{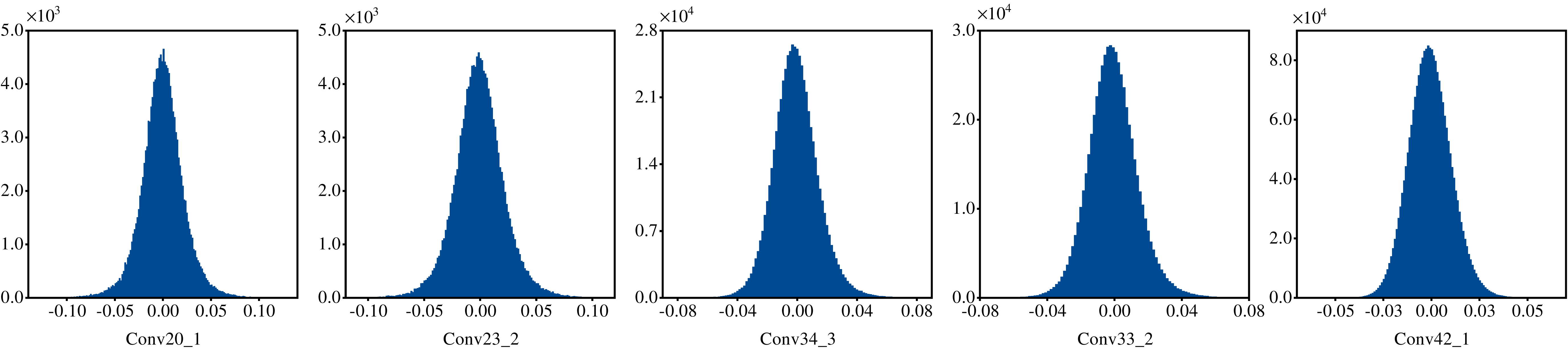}
\end{center}
\caption{\label{distribution}
Histograms of weights at different layers of ResNet-50, which have approximate zero means. Conv20\_1 denotes the 1st filter in the 20th convolutional layer and the same with others. Note that similar statistical result can be observed in other convolutional networks as well.
}
\end{figure*}

Optimization-based pruning leverages layer-wise optimization to minimize the reconstruction error between the full model and the pruned model. He~
\emph{et al.}~\cite{he2017channel} presented a LASSO-based filter selection strategy to identify representative filters and a least square reconstruction error to reconstruct the outputs. These two steps are iteratively executed until convergence. In \cite{luo2017thinet}, Luo~\emph{et al}. reconstructed the statistics information from the next layer to guide the importance evaluation of filters from the current layer. A set of training samples is used to deduce a closed-form solution.

The proposed FilterSketch can be grouped into optimization-based pruning but differs from \cite{he2017channel,luo2017thinet} in two aspects:
First, it preserves the second-order information of pre-trained weights, leading to quick accuracy recovery without the requirement of training from scratch or layer-wise fine-tuning. Second, it can be formulated as the matrix sketch problem and solved by the off-the-shelf Frequent Direction (FD) method, leading to a several-orders-of-magnitude reduction of time consumption without introducing data-driven and/or iterative optimization procedure. We note that \cite{kasiviswanathan2018network} conducts a complex tensor sketch for network approximation. Differently, our FilterSketch uses the efficient FD algorithm for the goal of information preserving. The network pruning and network approximation can be combined to further reduce the network size, which will be our future work.

Note that,~\cite{lecun1990optimal,hassibi1992second,dong2017learning,peng2019collaborative} also exploit the second order information for the network pruning. Nevertheless, our second-order information fundamentally differs from~\cite{lecun1990optimal,hassibi1992second,dong2017learning,peng2019collaborative}. First, the second-order information of~\cite{lecun1990optimal,hassibi1992second,dong2017learning,peng2019collaborative} lies in the second-order derivatives while our FilterSketch lies in the second-order covariance. Second, the second-order derivatives in~\cite{lecun1990optimal,hassibi1992second,dong2017learning,peng2019collaborative} are used as a measure to identify unimportant weights/channels while our FilterSketch aims to preserve the covariance information in the preserved weights. Third,~\cite{lecun1990optimal,hassibi1992second,dong2017learning,peng2019collaborative} involve heavy computation for constructing Hessian Matrix while our FilterSketch is implemented in a computationally cheap manner via the Frequent Direction.

\section{The Proposed Approach}\label{proposed_approach}
\subsection{Notations}\label{notations}
We start with notation definitions. Given a pre-trained CNN model $F$, which contains $L$ convolutional layers, and a set of filters $W = \{ W^i\}_{i=1}^L$ with $W^i = \{W^i_j\}_{j=1}^{c_i} \in \mathbb{R}^{d_i \times c_i}$ and $d_i = c_{i-1} \times h_i \times w_i$, where $c_i$, $h_i$ and $w_i$ respectively denote the channel number, filter height and width of the $i$-th layer. $W^i$ is the filter set for the $i$-th layer and $W_j^i$ is the $j$-th filter in the $i$-th layer.

The goal is to search for the pruned model $\mathcal{F}$, a set of transformed filters ${\Omega} = \{ {\Omega}^i \}_{i=1}^L$ with ${\Omega}^i = \{ {\Omega}_j^i \}_{j=1}^{\tilde{c}_i } \in \mathbb{R}^{d_i \times \tilde{c}_i}$ and $\tilde{c}_i = \lfloor p_i \cdot c_i \rceil$, where $p_i$ is the pruning rate for the $i$-th layer and $0 < p_i \le 1$. $\lfloor \cdot \rceil$ rounds the input to its nearest integer.

To learn ${\Omega}^i$ for each layer, predominant approaches are often divided into three streams:
(1) Retraining CNNs from scratch by imposing human-designed regularizations into the training loss \cite{huang2018data,lin2019towards}.
(2) Measuring the importance of filters via an intrinsic property of CNNs \cite{li2017pruning,yu2018nisp}.
(3) Minimizing the reconstruction error \cite{he2017channel,luo2017thinet} for pruning optimization.
Nevertheless, these methods solely consider the first-order statistics, while missing the covariance information.

\subsection{Information Preserving}\label{information_preserving}
In this study, we devise a novel second-order covariance preserving scheme, which provides a good warm-up for fine-tuning the pruned network. 
Different from existing works~\cite{kim2018bilinear,zhang2019discriminative,ren2019heterogeneous} where the covariance statistics of feature maps are calculated, we aim to preserve the covariance information of filters.

For each $W^i \in \mathbb{R}^{d_i \times c_i}$, our second-order preserving scheme aims to find a filter matrix ${\Omega}^i \in \mathbb{R}^{d_i \times \tilde{c}_i}$, which contains only $\tilde{c}_i \le c_i$ columns but preserves sufficient covariance information of $W^i$, as:
\begin{align}\label{covariance}
\Sigma_{W^i} \approx \Sigma_{{\Omega}^i},
\end{align}
where $\Sigma_{W^i}$ and $\Sigma_{{\Omega}^i}$ respectively denote the covariance matrices of $W^i$ and ${\Omega}^i$  and are defined as:
\begin{align}
    & \Sigma_{W^i} = (W^i - \bar{W}^i)(W^i - \bar{W}^i)^T, \label{1} \\ 
    & \Sigma_{{\Omega}^i} = ({\Omega}^i - {\bar{\Omega}}^i)({\Omega}^i - {\bar{\Omega}}^i)^T, \label{2}
\end{align}
where $\bar{W}^i = \frac{1}{c_i}\sum_{j=1}^{c_i}{W^i_j}$ and ${\bar{\Omega}}^i = \frac{1}{\tilde{c}_i}\sum_{j=1}^{\tilde{c}_i}{{\Omega}^i_j}$ are the mean values of the filters in the $i$-th layer for the full model and pruned model, respectively.

The covariance $\Sigma_{W^i}$ can effectively measure the pairwise interactions between the pre-trained filters.
A key ingredient of FilterSketch is that it can well preserve the correlation information of $W^i$ in ${\Omega}^i$.
Through this, it yields a more expressive and informative ${\Omega}^i$ for fine-tuning, as validated in Sec.\,\ref{experiment}.
To preserve the covariance information in Eq.\,\ref{covariance}, we formulate the following objective function:
\begin{equation}\label{minimize_covariance}
\mathop{\arg\min}_{{\Omega}^i} \| \Sigma_{W^i} - \Sigma_{{\Omega}^i} \|_F,
\end{equation}
where $\| \cdot \|_F$ denotes the Frobenius norm. Based on Eq.\,\ref{1} and Eq.\,\ref{2}, Eq.\,\ref{minimize_covariance} is expanded as:
\begin{equation}\label{expand_covariance}
\begin{split}
\mathop{\arg\min}_{{\Omega}^i} & \big\| (W^i - \bar{W}^i)(W^i - \bar{W}^i)^T - ({\Omega}^i - {\bar{\Omega}}^i)({\Omega}^i - {\bar{\Omega}}^i)^T \big\|_F.
\end{split}
\end{equation}

We statistically observe that $\bar{W}^i \approx 0$. As illustrated in Fig.\,\ref{distribution}, the pre-trained weights intend to follow a zero-mean Gaussian-like weight distribution, which is also discussed in many previous works~\cite{zhang2018systematic,he2019simultaneously,franchi2020tradi,glorot2010understanding,he2015delving}. The potential reason behind might be that the network is often trained with zero-mean Gaussian distribution as the initialization. During training, the regularization effect of $\ell_2$-norm weight penalty confines weights to the bell-shape histogram, and thus prevents a drastic change of the initial Gaussian distribution. As such, the pre-trained weights still have the approximately zero-mean Gaussian-like histogram. Similarly, it is intuitive that a good pruned weight ${\Omega}^i$ satisfies that $\mathcal{\bar{W}}^i \approx 0$. Thus, Eq.\,\ref{expand_covariance} can be re-written as:
\begin{equation}\label{simply_covariance}
\mathop{\arg\min}_{{\Omega}^i}\big\| W^i(W^i)^T - {\Omega}^i({\Omega}^i)^T \big\|_F.
\end{equation}

%
%

Similar to \,\cite{he2017channel,luo2017thinet}, one can develop a series of optimization steps to minimize the reconstruction error of Eq.\,\ref{simply_covariance}.
However, the optimization procedure is typically based on data-driven and/or iterative methods \cite {he2017channel,luo2017thinet}, which inevitably introduces heavy computation cost.

\subsection{Tractability}\label{filter_sketching}
In this section, we show that Eq.\,\ref{simply_covariance} can be effectively and efficiently solved by the off-the-shelf matrix sketch method \cite{liberty2013simple}, which does not involve data-driven iterative optimization while maintaining the property of interest of $W^i$ in ${\Omega}^i$.

Specifically, a sketch of a matrix $W^i$ is a transformed matrix ${\Omega}^i$, which is smaller than $W^i$ but tracks an $\varepsilon$-approximation to the norm of $W^i$, as:
\begin{equation}\label{sketch}
\begin{split}
&\qquad {\Omega}^i({\Omega}^i)^T \preccurlyeq W^i(W^i)^T\text{, and}
\\& \big\| W^i(W^i)^T - {\Omega}^i({\Omega}^i)^T \big\|_F \le \varepsilon \| W^i \|_F^2.
\end{split}
\end{equation}

Several papers have been devoted to solving Eq.\,\ref{sketch}, including CUR decomposition \cite{drineas2003pass}, random projection \cite{sarlos2006improved}, and column sampling methods \cite{frieze2004fast}, which however still rely on iterative optimization.

The streaming-based Frequent Direction (FD) method by \cite{liberty2013simple} provides a promising direction to solve this problem, where each sample is passed forward only once without iterations, which is extremely efficient. We summarize it in Alg.\,\ref{alg1}.
A $d_i \times c_i$ data matrix $W^i$ and the sketched size $\tilde{c}_i$ are fed into FD.
Each column $W^i_j$ of matrix $W^i$ represents a sample.
Columns from $W^i$ will replace all zero-valued columns in ${\Omega}^i$, and half of the columns in the sketch will be emptied with two steps once ${\Omega}^i$ is fully fed with non-zero valued columns:
In the first step, the sketch is rotated (from right) with the SVD decomposition of ${\Omega}^i$ so that its columns are orthogonal and in descending magnitude order. In the SVD decomposition, $USV^T = {\Omega}^i$, $U^TU=V^TV=VV^T=I_{\tilde{c}_i}$, where $I_{\tilde{c}_i}$ is the $\tilde{c}_i \times \tilde{c}_i$ identity matrix,
$S$ is a non-negative diagonal matrix and $S_{11} \ge ... \ge S_{\tilde{c}_i\tilde{c}_i} \ge 0$.
In the second step, $S$ is shrunk so that half of its singular values are zeros.
Accordingly, the right half of the columns in $U\hat{S}$ (see line 7 of Alg.\,\ref{alg1} for $\hat{S}$) will be zeros.
The details of the method can be referred to \cite{liberty2013simple}.

\begin{algorithm}[!t]
\begin{algorithmic}[1]
\caption{\label{alg1}Frequent Direction \protect\cite{liberty2013simple}.}
\REQUIRE
    Matrix $W^i \in \mathbb{R}^{d_i \times c_i}$, sketched size $\tilde{c}_i$.
\ENSURE
    Sketched matrix ${\Omega}^i \in \mathbb{R}^{d_i \times \tilde{c}_i}$.\\
     
\STATE ${\Omega}^i \leftarrow$ all zeros matrix $\in \mathbb{R}^{d_i \times \tilde{c}_i}$.

\FOR {each column $W_j^i$ in $W^i$}
    \STATE Insert $W_j^i$  into a zero valued column of ${\Omega}^i$.
    \IF{${\Omega}^i$ has no zero valued columns}
        \STATE $[U, S, V] = SVD({\Omega}^i)$
        \STATE $\delta = s^2_{\frac{\tilde{c}_i}{2}}$. {\scriptsize \# The squared $(\frac{\tilde{c}_i}{2})$-th entry of \textit{S}.}
        \STATE $\hat{S} = \sqrt{\max(S^2 - I_{\tilde{c}_i}\delta, 0)}$. {\scriptsize \# Identity matrix $I_{\tilde{c}_i}$ $\in \mathbb{R}^{\tilde{c}_i \times \tilde{c}_i}$.}
        \STATE ${\Omega}^i = U\hat{S}$. {\scriptsize \# At least half columns of ${\Omega}^i$ are zeros.}
    \ENDIF
\ENDFOR
\end{algorithmic}
\end{algorithm}

It can be seen that the optimization of Eq.\,\ref{simply_covariance} is similar to the matrix sketch problem of Eq.\,\ref{sketch},
though the existence of the upper bound term $\varepsilon\|W^i\|_2^2$ does not necessarily result in optimal ${\Omega}^i$ for Eq.\,\ref{simply_covariance}.

Nevertheless, in what follows, we show that Alg.\,\ref{alg1} can provide a tight convergence bound to solve the sketch problem of Eq.\,\ref{sketch} while the learned ${\Omega}^i$ can serve as a good warm-up for fine-tuning the pruned model as demonstrated in Sec.\,\ref{experiment}.

\textbf{Corollary 1}.
If ${\Omega}^i$ is the sketch result of matrix $W^i$ with the sketched size $\tilde{c}_i$ by Alg.\,\ref{alg1}, then it holds:
\begin{equation}
\begin{split}
&\quad\, 0 \preccurlyeq {\Omega}^i({\Omega}^i)^T \preccurlyeq W^i(W^i)^T\text{, and}
\\& 
\big\| W^i(W^i)^T - {\Omega}^i({\Omega}^i)^T \big\|_F \le \frac{2}{\tilde{c}_i} \| W^i \|_F^2,
\end{split}
\end{equation}
\emph{i.e.}, $\varepsilon = \frac{2}{\tilde{c}_i}$.

The proof of Corollary 1 can be referred to \cite{liberty2013simple}. Accordingly, the convergence bound of FD is proportional to $\frac{1}{\tilde{c}_i}$.
Smaller $\tilde{c}_i$ causes more error, which is intuitive since smaller $\tilde{c}_i$ means more pruned filters.
Besides, the sketch time is up-bounded by $\mathcal{O}(d_ic_i\tilde{c}_i)$ \cite{liberty2013simple}.
Sec.\,\ref{l2_efficiency} shows that the sketch process requires less than 2 seconds with CPU, which verifies the efficiency of FilterSketch.

In our experiments, we find that a small portion of the elements in ${\Omega}^i$ are unordinary larger especially for a small value of $\tilde{c}_i$, which damages the accuracy performance as demonstrated in Tab.\,\ref{l2}. This is understandable since the error bound does exist in Corollary 1. We can conduct a second-round sketch for $\frac{W^i}{\| {\Omega}^i \|_F}$ to solve this problem of unstable numerical values (see below for analysis). However, in what follows, we show that the second-round sketch can be eliminated by simply normalizing ${\Omega}^i$.

\textbf{Theorem 1}. 
If ${\Omega}^i$ is the sketch result by applying Alg.\,\ref{alg1} to matrix $W^i$ with the sketched size $\tilde{c}_i$, then for any constant $\beta$, $\beta{\Omega}^i$ is the result by applying Alg.\,\ref{alg1} to matrix $\beta W^i$.

\textbf{Proof}.
We start with Line 5 of Alg.\,\ref{alg1}, which can be modified as:

Line 5: $[U, \beta S, V] = SVD(\beta{\Omega}^i)$.

Correspondingly, Line 6 to Line 8 can be modified as:

Line 6: ${\beta}^2 \delta = ({\beta s_{\frac{\tilde{c}_i}{2}}})^2$;

Line 7: $\beta\hat{S} = \sqrt{\max\big({{\beta}^2 S}^2 - I_{\tilde{c}_i}{\beta}^2\delta, 0\big)}$;

Line 8: $\beta{\Omega}^i = U \beta \hat{S}$.

Thus, the sketch of $\beta W^i$ results in $\beta{\Omega}^i$, which completes the proof. $\hfill\blacksquare$

By setting $\beta = \frac{1}{\| {\Omega}^i \|_F}$, we can see that the sketch result of $\frac{W^i}{\| {\Omega}^i \|_F}$ is equal to $\frac{{\Omega}^i}{\|{\Omega}^i\|_F}$. Note that $\beta = \frac{1}{\| {\Omega}^i \|_F} < 1$ is always satisfied from our extensive experiments. Thus, the unordinary larger elements in ${\Omega^i}$ can be re-scaled to smaller ones. Finally, $\frac{{\Omega}^i}{\|{\Omega}^i\|_F}$ is fed to the slimmed network for fine-tuning\footnote{To sketch $\frac{W^i}{\| W^i \|_F}$ might be an alternative, the result of which is $\frac{{\Omega}^i}{\| W^i\|_F}$. However, we observe that $\| W^i \|_F$ is much larger than $\| \Omega \|_F$ and thus most elements in $\frac{{\Omega}^i}{\| W^i\|_F}$ are very close to zero, which is infeasible.}.
%

\begin{algorithm}[!t]
\begin{algorithmic}[1]
\caption{\label{alg2} FilterSketch Algorithm.}
\REQUIRE
    Pre-trained model $F$ with filter set $W = \{ W^i \}_{i=1}^L$.
\ENSURE
    Pruned model $\mathcal{F}$ with filter set ${\Omega} = \{ {\Omega}^i \}_{i=1}^L$ and per-layer pruning rate set $\{p_i\}_{i=1}^L$.\\
    
\FOR {$ i = 1 \rightarrow L$}
    \STATE Compute the sketched size by $\tilde{c}_i = \lfloor p_i \cdot c_i \rceil$.
    \STATE Obtain ${\Omega}^i$ via sketching $W^i$ by Alg.\,\ref{alg1}.
    \STATE Normalize ${\Omega}^i$ via $\ell_2$-norm.
\ENDFOR

\STATE Initialize $\mathcal{F}$ with the sketched filter set ${\Omega}$.

\FOR {$ t = 1 \rightarrow T$}
    \STATE Fine-tune the pruned model $\mathcal{F}$.
\ENDFOR

\STATE Return $\mathcal{F}$ with the fine-tuned filter set ${\Omega}$.
\end{algorithmic}
\end{algorithm}

We outline FilterSketch in Alg.\,\ref{alg2}. It can be seen that, compared with existing methods, FilterSketch stands out in that it is deterministic, simple to implement, and also very fast (see Tab.\,\ref{efficiency} later).

\section{Experiments}\label{experiment}
To show the effectiveness and efficiency of FilterSketch, we have conducted extensive experiments on image classification.
Representative compact-designed networks, including GoogLeNet \cite{szegedy2015going} and ResNet-50/56/110 \cite{he2016deep}, are chosen to compress. 
We report the performance of FilterSketch on CIFAR-10 \cite{krizhevsky2009learning} and ILSVRC-2012 \cite{russakovsky2015imagenet}, and compare it to state-of-the-arts (SOTAs) including regularization-based pruning \cite{huang2018data,lin2019towards}, property-based pruning \cite{li2017pruning,yu2018nisp,lin2020hrank}, and optimization-based pruning \cite{he2017channel,luo2017thinet}. Besides, we also conduct sub-sampling of the pre-trained weights for fine-tuning (denoted as Random) to show the advantage of considering the second-order information.

\begin{table*}[!t]
\caption{Results of GoogleNet and ResNet-56/110 on CIFAR-10.}
\centering
\begin{tabular}{ccccc}
\toprule
Model  &Top1-acc\  &$\uparrow\downarrow$ &FLOPs (Pruning Rate)      &Parameters (Pruning Rate)\\
\midrule
GoogLeNet (Base) &95.05\% &0.00\% &1.52B (0.0\%)  &6.15M (0.0\%) \\
L1 \cite{li2017pruning}  &94.54\%  &0.51\%$\downarrow$  &1.02B (32.9\%)   &3.51M (42.9\%) \\
GAL-0.05 \cite{lin2019towards}  &93.93\%  & 1.12\%$\downarrow$&0.94B (38.2\%)    &3.12M (49.3\%)  \\
HRank \cite{lin2020hrank} &94.53\%   &   0.52\%$\downarrow$&0.69B (54.9\%)    &2.74M (55.4\%) \\
Random  &93.23\% & 1.82\%$\downarrow$&0.59B (61.1\%) &2.61M (57.6\%)   \\
\textbf{FilterSketch} &\textbf{94.88\%} & \textbf{0.17\%$\downarrow$}&\textbf{0.59B (61.1\%)} &\textbf{2.61M (57.6\%)} \\
\bottomrule
ResNet-56 (Base) &93.26\% &0.00\%  &125.49M (0.0\%)  &0.85M (0.0\%)     \\
L1 \cite{li2017pruning}  &93.06\% &   0.20\%$\downarrow$&90.90M (27.6\%)   &0.73M (14.1\%) \\
He \emph{et al}. \cite{he2017channel}   &90.80\% & 2.46\%$\downarrow$&62.00M (50.6\%)     &-    \\
HRank~\cite{lin2020hrank} &93.52\% &0.26\%$\uparrow$ &88.72M (29.3\%)&0.71M (16.8\%)\\
\textbf{FilterSketch} &\textbf{93.65\%}&\textbf{0.39\%$\uparrow$}&\textbf{88.05M (30.43\%)}&\textbf{0.68M (20.6\%)} \\
NISP \cite{yu2018nisp} &93.01\%  & 0.25\%$\downarrow$&81.00M (35.5\%)    &0.49M (42.4\%)  \\
GAL-0.6 \cite{lin2019towards} &92.98\%  & 0.28\%$\downarrow$&78.30M (37.6\%)   &0.75M (11.8\%)  \\
Random     &90.55\% &  2.71\%$\downarrow$&73.36M (41.5\%) &0.50M (41.2\%) \\
\textbf{FilterSketch} &\textbf{93.19\%} &\textbf{0.07\%$\downarrow$}&\textbf{73.36M (41.5\%)} &\textbf{0.50M (41.2\%)} \\
HRank~\cite{lin2020hrank} &90.72\%&2.54\%$\downarrow$&32.52M (74.1\%)&0.27M (68.1\%)\\
\textbf{FilterSketch} &\textbf{91.20\%}&\textbf{2.06\%$\downarrow$}&\textbf{32.47M (74.4\%)}&\textbf{0.24M (71.8\%)} \\
\midrule
ResNet-110 (Base) &93.50\%  &0.00\% &252.89M (0.0\%) &1.72M (0.0\%)   \\
L1 \cite{li2017pruning} &93.30\%  &    0.20\%$\downarrow$&155.00M (38.7\%)  &1.16M (32.6\%) \\
GAL-0.5 \cite{lin2019towards}  &92.55\%  &   0.95\%$\downarrow$&130.20M (48.5\%)  &0.95M (44.8\%)\\
HRank \cite{lin2020hrank}  &93.36\%  & 0.14\%$\downarrow$&105.70M (58.2\%)  &0.70M (59.2\%) \\
Random    &89.88\% & 3.62\%$\downarrow$&92.84M (63.3\%) &0.69M (59.9\%) \\
\textbf{FilterSketch} &\textbf{93.44\%} &\textbf{0.06\%$\downarrow$}&\textbf{92.84M (63.3\%)} &\textbf{0.69M (59.9\%)} \\
\bottomrule
\end{tabular}
\label{cifar}
\end{table*}

\subsection{Implementation Details}
\textbf{Training Strategy}. 
We use the Stochastic Gradient Descent (SGD) for fine-tuning with the Nesterov momentum 0.9 and the batch size is set to 256. For CIFAR-10, the weight decay is set to 5e-3 and we fine-tune the network for 150 epochs with an initial learning rate of 0.01, which is then divided by 10 every 50 epochs. For ILSVRC-2012, the weight decay is set to 5e-4 and 90 epochs are given to fine-tune the network. The learning rate is initially set to 0.1, and divided by 10 every 30 epochs. 

For all methods, we apply the standard data augmentation provided by the official Pytorch including random crop and horizontal flip. To stress, other techniques for image augmentation, such as lightening and color jitter, can be applied to further improve the performance as done in the implementations of~\cite{yu2019autoslim,liu2019metapruning,ding2019approximated}, or even the cosine learning rate \cite{dong2019network}, can also be applied to further improve the accuracy performance. We do not consider these since we aim to show the performance of pruning algorithms themselves. We provide our codes in the supplementary material.

\textbf{Performance Metric}.
Parameter amount and FLOPs (floating-point operations) are used as the metrics, which respectively denote the storage and computation cost. We also report the pruning rate (PR) of parameters and FLOPs. For CIFAR-10, top-1 accuracy are provided.
For ILSVRC-2012, both top-1 and top-5 accuracies are reported.

\subsection{Results on CIFAR-10}\label{result_cifar}

We evaluate the performance of FilterSketch on CIFAR-10 with popular networks, including GoogLeNet, ResNet-56 and ResNet-110.
For GoogLeNet, we make the final output class number the same as the number of categories on CIFAR-10.

\textbf{GoogLeNet}. 
As can be seen from Tab.\,\ref{cifar}, FilterSketch outperforms the SOTA methods in both accuracy retaining and model complexity reductions. Specifically, 61.1\% of the FLOPs are reduced and 57.6\% of the parameters are removed, achieving a significantly higher compression rate than GAL-0.6 and HRank. Besides, FilterSketch also maintains a comparable top-1 accuracy, even better than L1, which obtains a much less complexity reduction.

\textbf{ResNet-56}.
Results for ResNet-56 are presented in Tab.\,\ref{cifar}, where FilterSketch removes around 41.5\% of the FLOPs and parameters while keeping the top-1 accuracy at 93.19\%. Compared to 93.26\% by the pre-trained model, the accuracy drop is negligible. Compared with L1, FilterSketch shows an overwhelming superiority. Though NISP obtains 1\% more parameters reduction than FilterSketch, it takes more computation in the convolutional layers with a lower top-1 accuracy. Moreover, in comparison with HRank under similar reductions of FLOPs and parameters, our FilterSketch results in a higher top-1 accuracy, well demonstrating the effectivness of the sketched weights to recover the accuracy performance.
%

\begin{table*}[!t]
\caption{Results of ResNet-50 on ILSVRC-2012.}
\begin{tabular}{ccccccc}
\toprule
Model  &Top1-acc\% &$\uparrow\downarrow$  &Top5-acc\% &$\uparrow\downarrow$  &FLOPs (Pruning Rate)     &Parameters (Pruning Rate)\\
\midrule
ResNet-50 (Base) \cite{luo2017thinet}  &76.13\% &0.00\%   &92.86\% &0.00\%   &4.09B (0.0\%)  &25.50M (0.0\%)    \\
SSS-32~\cite{huang2018data} &74.18\% &   1.95\%$\downarrow$&91.91\% &  0.95\%$\downarrow$&2.82B (31.1\%)  &18.60M (27.1\%)  \\
He \emph{et al}. \cite{he2017channel}&72.30\% & 3.83\%$\downarrow$&90.80\% & 2.06\%$\downarrow$&2.73B (33.3\%)   & -      \\
FilterSketch-0.7  &\textbf{75.22\%}  &\textbf{0.91\%$\downarrow$} &\textbf{92.41\%}  &\textbf{0.45\%$\downarrow$}&\textbf{2.64B (35.5\%)}  &16.95M (33.5\%) \\
GAL-0.5 \cite{lin2019towards}&71.95\% & 4.18\%$\downarrow$&90.94\%  &  1.92\%$\downarrow$&2.33B (43.0\%) &21.20M (16.9\%)  \\
SSS-26 \cite{huang2018data}  &71.82\%   & 4.31\%$\downarrow$&90.79\% & 2.07\%$\downarrow$&2.33B (43.0\%)  &15.60M (38.8\%)  \\
\textbf{FilterSketch-0.6} &\textbf{74.68\%} &\textbf{1.45\%$\downarrow$}&\textbf{92.17\%} &\textbf{0.69\%$\downarrow$}&\textbf{2.23B (45.5\%)}  &\textbf{14.53M (43.0\%)} \\
HRank  \cite{lin2020hrank}  &71.98\% & 4.15\%$\downarrow$&91.01\% & 1.85\%$\downarrow$&1.55B (62.1\%) &13.77M (46.0\%) \\
GAL-0.5-joint \cite{lin2019towards}&71.80\% & 4.33\%$\downarrow$&90.82\% & 2.04\%$\downarrow$&1.84B (55.0\%)  &19.31M (24.3\%)  \\
%
%
ThiNet-50 \cite{luo2017thinet}   &71.01\%&   5.12\%$\downarrow$&90.02\% & 2.84\%$\downarrow$&1.71B (58.2\%)   &12.28M (51.8\%)   \\
GAL-1 \cite{lin2019towards} &69.88\%  &  6.25\%$\downarrow$&89.75\% & 3.11\%$\downarrow$&1.58B (61.4\%)  &14.67M (42.5\%)  \\
\textbf{FilterSketch-0.4}  &\textbf{73.04\%}  &\textbf{3.09\%$\downarrow$}&\textbf{91.18\%}  & \textbf{1.68\%$\downarrow$}&\textbf{1.51B (63.1\%)}  &\textbf{10.40M (59.2\%)} \\
ThiNet-50 \cite{luo2017thinet}  &68.42\% & 6.82\%$\downarrow$&88.30\% & 4.56\%$\downarrow$&1.10B (73.1\%)  &8.66M (66.0\%)     \\
GAL-1-joint  \cite{lin2019towards}  &69.31\%  &  6.82\%$\downarrow$&89.12\%  &  3.74\%$\downarrow$&1.11B (72.9\%)  &10.21M (60.0\%)  \\
HRank \cite{lin2020hrank} &68.10\% & 8.03\%$\downarrow$&89.58\% & 3.28\%$\downarrow$& 0.98B (76.0\%) &8.27M (67.6\%) \\
\textbf{FilterSketch-0.2} &\textbf{69.43\%}  &\textbf{6.70\%$\downarrow$}&\textbf{89.23\%}   & \textbf{3.63\%$\downarrow$}&\textbf{0.93B (77.3\%)}  &\textbf{7.18M (71.8\%)} \\
%
%
\bottomrule
\end{tabular}
\centering
\label{resnet50}
\end{table*}

\begin{figure}[!t]
\begin{center}
\includegraphics[height=0.65\linewidth]{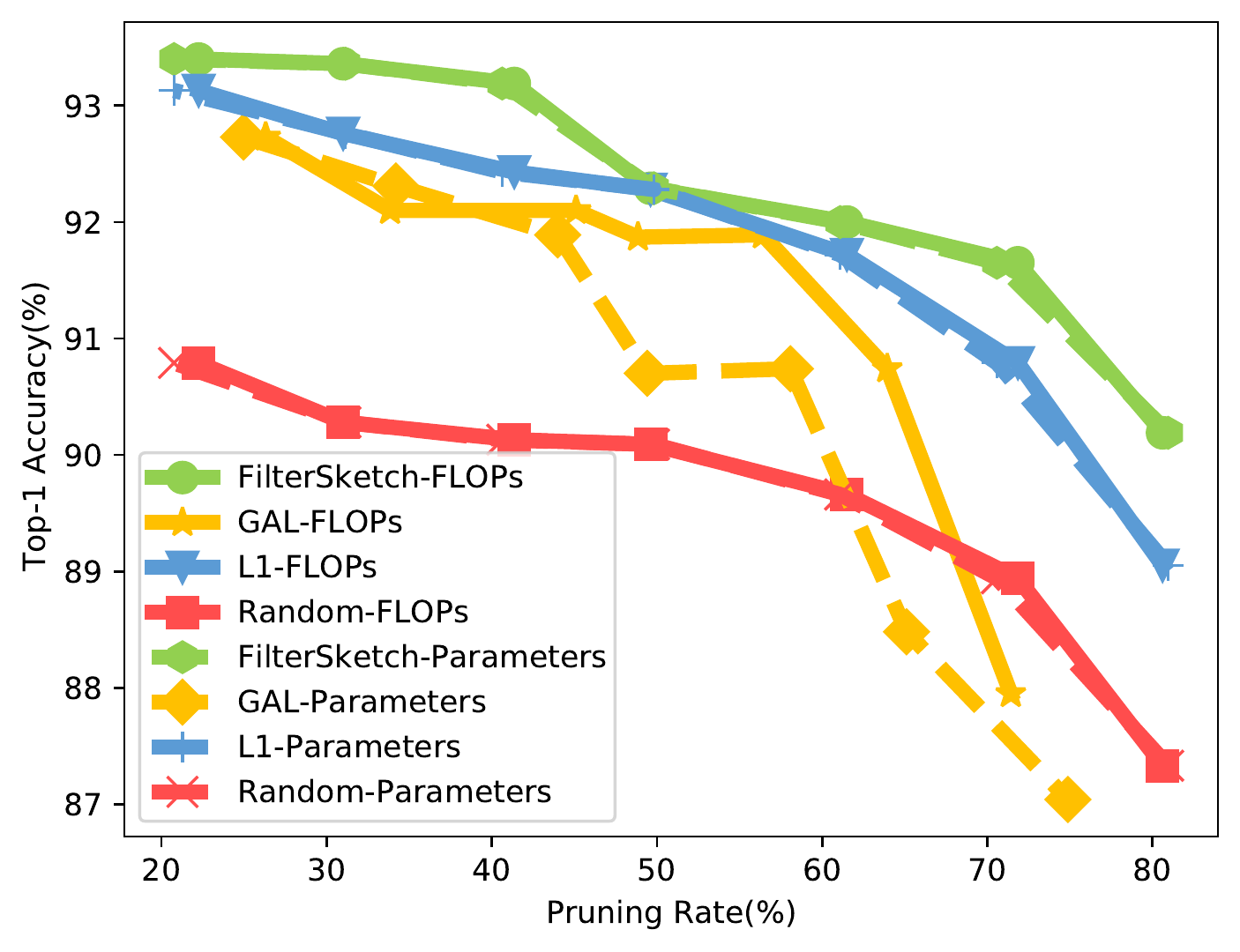}
\end{center}
\caption{\label{vary_rate}FLOPs and parameter comparison among GAL \cite{lin2019towards}, L1~\cite{li2017pruning}, Random, and our FilterSketch under different compression rates. ResNet-56 is compressed and Top-1 accuracy is reported.
}
\end{figure}

\textbf{ResNet-110}.
Tab.\,\ref{cifar} also displays the pruning results for ResNet-110. FilterSketch reduces the FLOPs of ResNet-110 by an impressive factor of 63.3\%, and the parameters by 59.9\%, while maintaining an accuracy of 93.44\%. FilterSketch significantly outperforms these SOTAs, showing that it can greatly facilitate the ResNet model, a popular backbone for object detection and semantic segmentation, to be deployed on mobile devices.

In Tab.\,\ref{cifar}, we also display the performance of randomly sub-sampling of filter weights (Random) given the same pruning rates as with FilterSketch. As seen, Random suffers great accuracy degradation in comparison with FilterSketch. In contrast, FilterSketch considers all the information in the pre-trained weights, which provides a more informative warm-up for fine-tuning the pruned model.

In Fig.\,\ref{vary_rate}, we further compare the Top-1 accuracies of the compressed models by GAL~\cite{lin2019towards}, L1~\cite{li2017pruning}, Random, and our FilterSketch under different compression rates using ResNet-56. As shown in the figure, our method outperforms the compared methods easily. Especially, for large pruning rates (> 60\%), both L1 and GAL suffer an extreme accuracy drop while FilterSketch maintains a relatively stable performance, which stresses the importance of information preserving in network pruning again.

\subsection{Results on ILSVRC-2012}\label{result_imagenet}
In Tab.\,\ref{resnet50}, we show the results for ResNet-50 on ILSVRC-2012 and compare FilterSketch to many SOTAs.
We display different pruning rates for FilterSketch and compare top-1 and top-5 accuracies.
%
%
For convenience, we use FilterSketch-$\alpha$ to denote the sketch rate $\alpha$ (\emph{i.e.}, $\alpha = \frac{\tilde{c}}{c}$) for FilterSketch.
Smaller $\alpha$ leads to a higher compression rate.

As shown in Tab.\,\ref{resnet50}, with similar or better reductions of FLOPs and parameters, FilterSketch demonstrates its great advantages in retaining the accuracy in comparisons with the SOTAs.
For example, FilterSketch-0.6 obtains 74.68\% top-1 and 92.17\% top-5 accuracies, significantly better than GAL-0.5 and SSS-26.
Another observation is that, with similar or more FLOPs reduction, FilterSketch also removes more parameters.
Hence, FilterSketch is especially suitable for network compression.

\begin{figure}[!t]
\begin{center}
\includegraphics[height=0.55\linewidth]{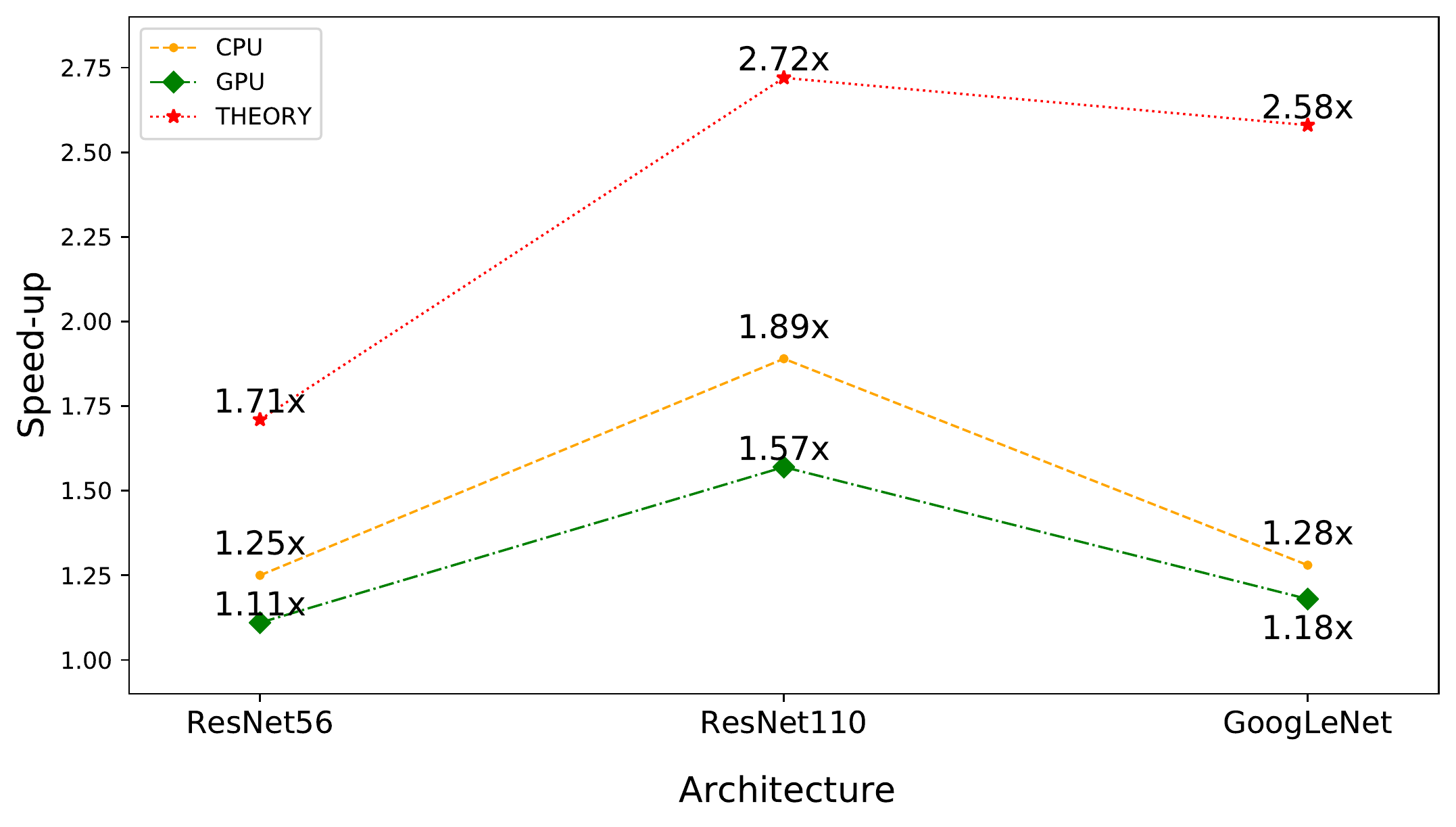}
\end{center}
\caption{\label{speedup}Speedups corresponding to CPU (Intel(R) Xeon(R) CPU E5-2620 v4 @2.10GHz) and GPU (GTX-1080TI) over the different CNNs with a batch size 256 on CIFAR-10. 
}
\end{figure}

\subsection{Practical Speedup}
The practical speedup for pruned CNNs depends on many factors, \emph{e.g.}, FLOPs reduction percentage, the number of CPU/GPU cores available and I/O delay of data swap, \emph{etc}.

We test the speedup of pruned models by FilterSketch in Tab.\,\ref{cifar} with CPU and GPU, and present the results in Fig.\,\ref{speedup}. Compared with the theoretical speedups of 1.71$\times$, 2.72$\times$ and 2.58$\times$ for ResNet-56, ResNet-110 and GoogLeNet, respectively. FilterSketch gains 1.11$\times$, 1.57$\times$ and 1.18$\times$ practical speedups with GPU while 1.25$\times$, 1.89$\times$ and 1.28$\times$ with CPU are obtained.

\subsection{Normalization Influence and Optimization Efficiency}\label{l2_efficiency}%
To measure the effectiveness of the sketch with the Frobenius normalization, we compare our FilterSketch models given in Tab.\,\ref{cifar} and Tab.\,\ref{resnet50} (FilterSketch-0.4) with corresponding models but excluding the Frobenius normalization. As shown in Tab.\,\ref{l2}, the former (the third column) obtains a better accuracy than the latter (the second column), which verifies the analysis in Sec.\,\ref{filter_sketching} that the sketch with the Frobenius normalization can effectively solve the problem of unstable numerical values after sketch. As for the sketch efficiency, we again compare these four models in Tab.\,\ref{l2} with two optimization-based methods \cite{luo2017thinet,he2017channel}. The results in Tab.\,\ref{efficiency} show that the time cost in the sketch process is little. Even with wider GoogLeNet and deeper ResNet-110, the sketches consume less than 2 seconds, which are several orders of magnitude faster than the other methods that cost many hours, or even days.

\begin{table}[!t]
\caption{\label{l2}Performance comparisons between sketches with and without the normalization.}
\centering
\begin{tabular}{c|c|c}
\hline
          & Sketch accuracy & Sketch+norm accuracy   \\ \hline
GoogLeNet    &94.54\%   &94.88\% \\ \hline
ResNet-56    &92.81\%   &93.19\%   \\ \hline
ResNet-110   &93.01\%   &93.44\%     \\ \hline
ResNet-50    &72.84\% / 91.01\%  &73.04\% / 91.18\%    \\ \hline
\end{tabular}
\end{table}

\begin{table}[!t]
\caption{\label{efficiency}Optimization efficiency (CPU) among ThiNet \protect\cite{luo2017thinet}, CP \protect\cite{he2017channel} and FilterSketch.}
\centering
\begin{tabular}{c|c|c|c}
\hline
             & ThiNet  & CP  &\textbf{FilterSketch} \\ \hline
GoogLeNet    &183585.41s   &2008.72s   &1.85s\\ \hline
ResNet-56    &24422.73s  &536.55s  &0.09s  \\ \hline
ResNet-110   &63695.89s  &961.71s  &1.06s  \\ \hline
ResNet-50    &4130102.59s &205117.20s  &1.41s    \\ \hline
\end{tabular}
\end{table}

\section{Conclusion}\label{conclusion}
We have proposed a novel approach, termed FilterSketch, for structured network pruning. Instead of simply discarding unimportant filters, FilterSketch preserves the second-order information of the pre-trained model, through which the accuracy is well maintained. We have further proposed to obtain the information preserving constraint by utilizing the off-the-shelf matrix sketch method, based on which the requirement of training from scratch or iterative optimization can be eliminated, and the pruning complexity is significantly reduced. Extensive experiments have demonstrated the superiorities of FilterSketch over the state-of-the-arts. As the first attempt on weight information preserving, FilterSketch provides a fresh insight for the network pruning problem. Nevertheless, our FilterSketch is built on the fact that the filter weights have approximate zero mean in each layer of the convolutional neural network. Such a requirement might not be satisfied in other networks, \emph{e.g.}, multi-layer perceptron. More effort will be made to solve this issue in our future work.
%

\ifCLASSOPTIONcaptionsoff
  \newpage
\fi

%
%
\bibliographystyle{IEEEtran}
\bibliography{reference}

\begin{IEEEbiography}[{\includegraphics[width=1in,height=1.25in,clip,keepaspectratio]{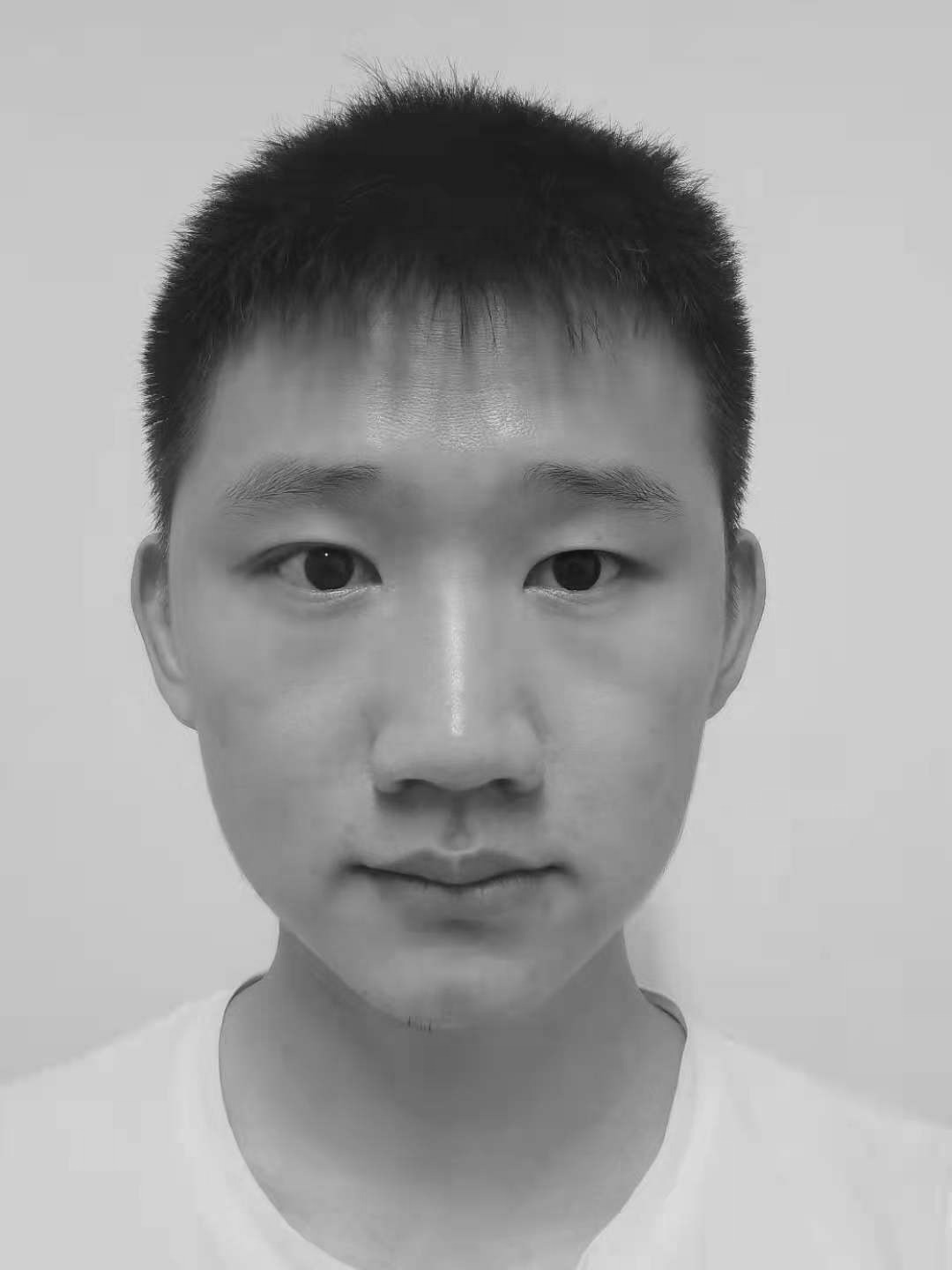}}]{Mingbao Lin}
is currently pursuing the Ph.D degree with Xiamen University, China. He has published over ten papers as the first author in international journals and conferences, including IEEE TPAMI, IJCV, IEEE TIP, IEEE TNNLS, IEEE CVPR, NeuriPS, AAAI, IJCAI, ACM MM and so on. His current research interest includes network compression \& acceleration, and information retrieval.
\end{IEEEbiography}

\begin{IEEEbiography}[{\includegraphics[width=1in,height=1.25in,clip,keepaspectratio]{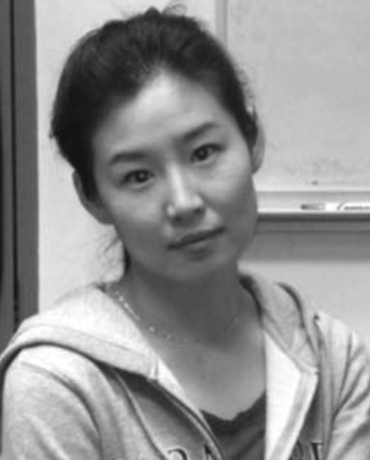}}]{Liujuan Cao}
received the B.S., M.S., and Ph.D degrees from the School of Computer Science and Technology, Harbin Engineering University. She is currently an associate professor at Xiamen University. Her research interest mainly focuses on computer vision and pattern recognition. She has authored over 40 papers in top and major tired journals and conferences, including CVPR, TIP, etc. She is the Financial Chair of the IEEE MMSP 2015, the Workshop Chair of the ACM ICIMCS 2016, and the Local Chair of the Visual and Learning Seminar 2017.
\end{IEEEbiography}

\begin{IEEEbiography}[{\includegraphics[width=1in,height=1.25in,clip,keepaspectratio]{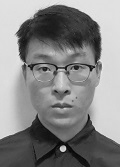}}]{Shaojie Li}
studied for his B.S. degrees in FuZhou University, China, in 2019. He is currently trying to pursue a M.S. degree in Xiamen University, China. His research interests include model compression and computer vision.
\end{IEEEbiography}

\begin{IEEEbiography}[{\includegraphics[width=1in,height=1.25in,clip,keepaspectratio]{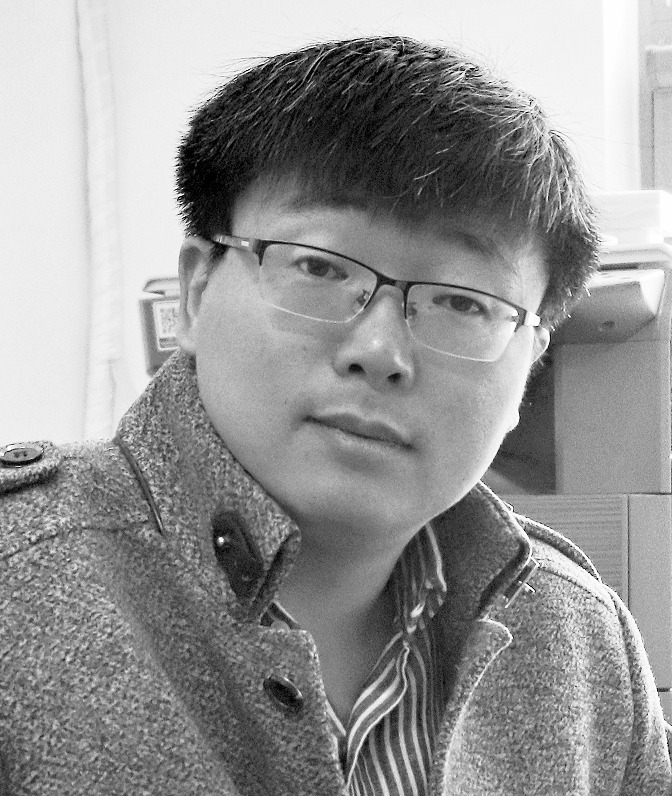}}]{Qixiang Ye} (Senior Member, IEEE) received the B.S. and M.S. degrees in mechanical and electrical engineering from Harbin Institute of Technology, China, in 1999 and 2001, respectively, and the Ph.D. degree from the Institute of Computing Technology, Chinese Academy of Sciences in 2006. He has been a professor with the University of Chinese Academy of Sciences since 2009, and was a visiting assistant professor with the Institute of Advanced Computer Studies (UMIACS), University of Maryland, College Park until 2013. His research interests include image processing, object detection and machine learning. He has published more than 100 papers in refereed conferences and journals including IEEE CVPR, ICCV, ECCV, NeurIPS, TNNLS, and PAMI.
\end{IEEEbiography}

\begin{IEEEbiography}[{\includegraphics[width=1in,height=1.25in,clip,keepaspectratio]{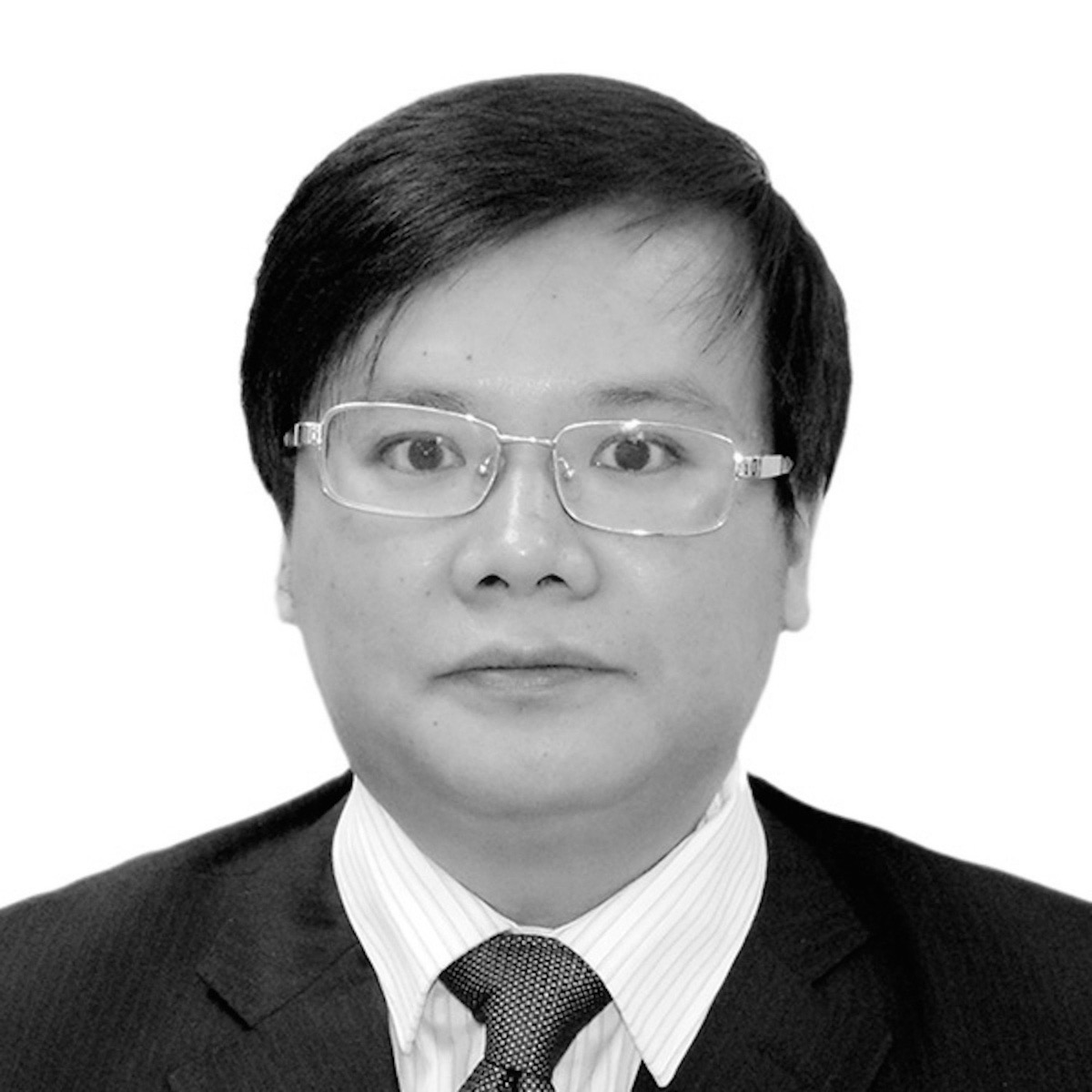}}]{Yonghong Tian} (Senior Member, IEEE)
is currently a Boya Distinguished Professor with the Department of Computer Science and Technology, Peking University, China, and is also the deputy director of Artificial Intelligence Research Center, PengCheng Laboratory, Shenzhen, China. His research interests include neuromorphic vision, brain-inspired computation and multimedia big data. He is the author or coauthor of over 200 technical articles in refereed journals such as IEEE TPAMI/TNNLS/TIP/TMM/TCSVT/TKDE/TPDS, ACM CSUR/TOIS/TOMM and conferences such as NeurIPS/CVPR/ICCV/AAAI/ACMMM/WWW. Prof. Tian was/is an Associate Editor of IEEE TCSVT (2018.1-), IEEE TMM (2014.8-2018.8), IEEE Multimedia Mag. (2018.1-), and IEEE Access (2017.1-). He co-initiated IEEE Int’l Conf. on Multimedia Big Data (BigMM) and served as the TPC Co-chair of BigMM 2015, and aslo served as the Technical Program Co-chair of IEEE ICME 2015, IEEE ISM 2015 and IEEE MIPR 2018/2019, and General Co-chair of IEEE MIPR 2020 and ICME2021. He is the steering member of IEEE ICME (2018-) and IEEE BigMM (2015-), and is a TPC Member of more than ten conferences such as CVPR, ICCV, ACM KDD, AAAI, ACM MM and ECCV. He was the recipient of the Chinese National Science Foundation for Distinguished Young Scholars in 2018, two National Science and Technology Awards and three ministerial-level awards in China, and obtained the 2015 EURASIP Best Paper Award for Journal on Image and Video Processing, and the best paper award of IEEE BigMM 2018. He is a senior member of IEEE, CIE and CCF, a member of ACM.
\end{IEEEbiography}

\begin{IEEEbiography}[{\includegraphics[width=1in,height=1.25in,clip,keepaspectratio]{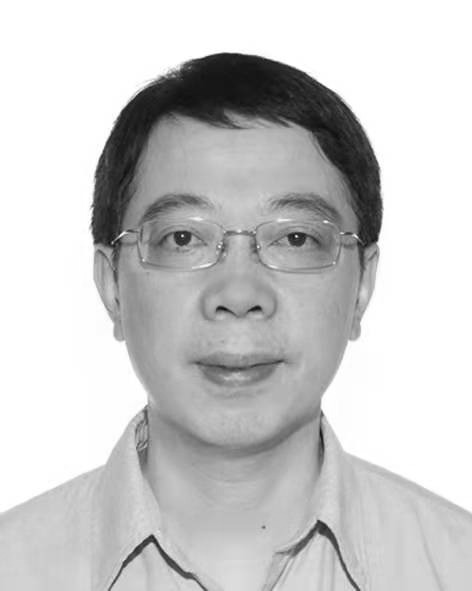}}]{Jianzhuang Liu} (Senior Member, IEEE) received the Ph.D. degree in computer vision from The Chinese University of Hong Kong, Hong Kong, in 1997. From 1998 to 2000, he was a Research Fellow with Nanyang Technological University, Singapore. From 2000 to 2012, he was a Postdoctoral Fellow, an Assistant Professor, and an Adjunct Associate Professor with The Chinese University of Hong Kong. In 2011, he joined the Shenzhen Institute of Advanced Technology, University of Chinese Academy of Sciences, Shenzhen, China, as a Professor. He is currently a Principal Researcher with Huawei Technologies Company Limited, Shenzhen, China. He has authored more than 150 papers. His research interests include computer vision, image processing, deep learning, and graphics.
\end{IEEEbiography}

\begin{IEEEbiography}[{\includegraphics[width=1in,height=1.25in,clip,keepaspectratio]{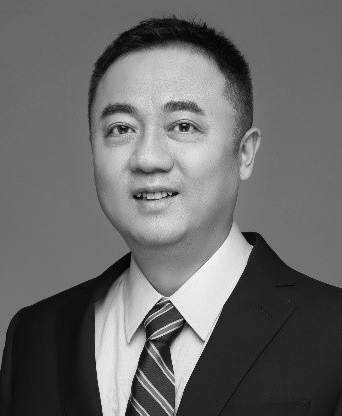}}]{Qi Tian} (Fellow, IEEE) is currently a Chief Scientist in Artificial Intelligence at Cloud BU, Huawei. From 2018-2020, he was the Chief Scientist in Computer Vision at Huawei Noah’s Ark Lab.  Before that he was a Full Professor in the Department of Computer Science, the University of Texas at San Antonio (UTSA) from 2002 to 2019. During 2008-2009, he took one-year Faculty Leave at Microsoft Research Asia (MSRA). 

Dr. Tian received his Ph.D. in ECE from University of Illinois at Urbana-Champaign (UIUC) and received his B.E. in Electronic Engineering from Tsinghua University and M.S. in ECE from Drexel University, respectively. Dr. Tian’s research interests include computer vision, multimedia information retrieval and machine learning and published 610+ refereed journal and conference papers. His Google citation is over 27900+ with H-index 81. He was the co-author of best papers including IEEE ICME 2019, ACM CIKM 2018, ACM ICMR 2015, PCM 2013, MMM 2013, ACM ICIMCS 2012, a Top 10\% Paper Award in MMSP 2011, a Student Contest Paper in ICASSP 2006, and co-author of a Best Paper/Student Paper Candidate in ACM Multimedia 2019, ICME 2015 and PCM 2007.

Dr. Tian received 2017 UTSA President’s Distinguished Award for Research Achievement, 2016 UTSA Innovation Award, 2014 Research Achievement Awards from College of Science, UTSA, 2010 Google Faculty Award, and 2010 ACM Service Award. He is the associate editor of IEEE TMM, IEEE TCSVT, ACM TOMM, MMSJ, and in the Editorial Board of Journal of Multimedia (JMM) and Journal of MVA.  Dr. Tian is the Guest Editor of IEEE TMM, Journal of CVIU, etc. Dr. Tian is a Fellow of IEEE.

\end{IEEEbiography}

\begin{IEEEbiography}[{\includegraphics[width=1in,height=1.25in,clip,keepaspectratio]{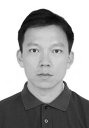}}]{Rongrong Ji}
(Senior Member, IEEE) is a Nanqiang Distinguished Professor at Xiamen University, the Deputy Director of the Office of Science and Technology at Xiamen University, and the Director of Media Analytics and Computing Lab. He was awarded as the National Science Foundation for Excellent Young Scholars (2014), the National Ten Thousand Plan for Young Top Talents (2017), and the National Science Foundation for Distinguished Young Scholars (2020). His research falls in the field of computer vision, multimedia analysis, and machine learning. He has published 50+ papers in ACM/IEEE Transactions, including TPAMI and IJCV, and 100+ full papers on top-tier conferences, such as CVPR and NeurIPS. His publications have got over 10K citations in Google Scholar. He was the recipient of the Best Paper Award of ACM Multimedia 2011. He has served as Area Chairs in top-tier conferences such as CVPR and ACM Multimedia. He is also an Advisory Member for Artificial Intelligence Construction in the Electronic Information Education Committee of the National Ministry of Education.
\end{IEEEbiography}

\end{document}